\documentclass[a4paper, conference]{IEEEtran}
\IEEEoverridecommandlockouts
\usepackage{abbrv}
\usepackage[dvipsnames]{xcolor}
\usepackage{url}
\usepackage{tikz}
\usetikzlibrary{positioning, arrows.meta, fit}
\usepackage{comment}
\usepackage{amsthm}
\usepackage{amsmath,amssymb,bbm}
\usepackage{colortbl}
\usepackage{epsfig}
\usepackage{caption}

\usepackage{soul}
\usepackage{orcidlink}
\usepackage[utf8]{inputenc}
\usepackage{listings}
\usepackage[T1]{fontenc}
\usepackage{booktabs}
\usepackage{amsfonts}
\usepackage{nicefrac}
\usepackage{makecell}
\usepackage{microtype}
\usepackage{threeparttable}
\usepackage{algorithm-mine}
\usepackage{numprint}
\usepackage{siunitx}
\usepackage{etoolbox}
\usepackage{cancel}
\newcommand{\ubold}{\fontseries{b}\selectfont}
\robustify\ubold
\usepackage{wrapfig}
\usepackage{algpseudocode}
\usepackage[skip=2pt,font=footnotesize,labelfont=footnotesize]{subcaption}
\usepackage{pifont}
\usepackage{lineno}
\usepackage{array,multirow,graphicx}
\usepackage{tabularx}
\usepackage{listings}
\usepackage[para]{footmisc}
\usepackage{lettrine}

\usepackage[accsupp]{axessibility}
\usepackage{accsupp}
\usepackage{pgfplots}
\usepackage[justification=centering]{caption}
\usepackage[T1]{fontenc}
\usepackage{float}
\usepackage{array}
\usepackage{amsmath}
\usepackage{changepage}
\pgfplotsset{compat=1.16}

\definecolor{cvprblue}{rgb}{0.21,0.49,0.74}

\definecolor{tabhighlight}{HTML}{e5e5e5}
\definecolor{lightblue}{rgb}{0.63, 0.79, 0.95}
\definecolor{babypink}{rgb}{0.96, 0.76, 0.76}
\definecolor{skyblue}{rgb}{0.53, 0.81, 0.92}
\definecolor{wheat}{rgb}{0.96, 0.87, 0.7}
\definecolor{delim}{RGB}{20,105,176}
\colorlet{punct}{red!60!black}
\colorlet{numb}{magenta!60!black}

\def\BibTeX{{\rm B\kern-.05em{\sc i\kern-.025em b}\kern-.08em
    T\kern-.1667em\lower.7ex\hbox{E}\kern-.125emX}}
\hypersetup{colorlinks,%
citecolor=green,%
filecolor=
linkcolor=%
urlcolor=green
}

\begin{document}

\title{SAGE-Yoga: Multi-Cue Learning for Yoga Pose Classification and Joint-Level Correction}

\makeatletter
\newcommand{\linebreakand}{%
  \end{@IEEEauthorhalign}
  \hfill\mbox{}\par
  \mbox{}\hfill\begin{@IEEEauthorhalign}
}
\makeatother

\author{
\IEEEauthorblockN{
Hung Le Chi,
Khanh Minh Huynh,
Long Nghia Tran Pham,
Tan Phuc Huynh, \\
Trong-Thuan Nguyen,
and Minh-Triet Tran
}
\IEEEauthorblockA{
\textit{University of Science, VNU-HCM, Ho Chi Minh City, Vietnam}\\
\textit{Vietnam National University, Ho Chi Minh City, Vietnam}\\
\texttt{
\{lchung2426, hkminh2420, tplnghia2416, htphuc2424\}@apcs.fitus.edu.vn
}\\
\texttt{
\{ntrthuan, tmtriet\}@hcmus.edu.vn
}
}
}
\maketitle

\begin{abstract}
Automated yoga analysis requires both accurate pose classification and interpretable feedback on pose execution. However, existing methods often rely on a single visual prediction, struggle to distinguish visually similar poses, and treat pose classification and correction as separate tasks. To address these limitations, we propose SAGE-Yoga, a unified coarse-to-fine framework for yoga pose classification and joint-level correction from a single RGB image. Inspired by how yoga instructors assess posture using multiple complementary cues, SAGE-Yoga first employs a bagging-based ensemble of complementary visual backbones to generate a ranked set of candidate pose classes. Additionally, a margin-based gating mechanism preserves confident visual predictions while invoking geometric verification only for ambiguous cases. Moreover, once the final pose class is determined, SAGE-Yoga retrieves a medoid reference pose and compares the observed joint angles with class-specific distributions to identify misaligned joints. Finally, these deviations are translated into actionable corrective feedback. Empirically, experiments on the Yoga-82 dataset show that the visual ensemble achieves 89.0\% Top-1 accuracy, while the complete framework improves performance to 90.7\% Top-1 accuracy and 90.1\% Macro-F1. These results demonstrate that combining complementary visual evidence with selective geometric verification improves fine-grained pose classification while enabling interpretable, joint-level correction.
\end{abstract}


\section{Introduction}\label{sec:intro}
\vspace{-2mm}
With the rapid growth of home-based fitness and wellness technologies, there is increasing demand for systems that provide accurate and interpretable guidance during unsupervised exercise. Specifically, yoga pose analysis is particularly challenging, as it requires precise pose recognition and meaningful assessment of execution quality. Therefore, automated yoga pose classification and correction~\cite{ref_article16, ref_article18, ref_article19} has become an important problem, aiming to predict pose categories and generate actionable, joint-level feedback for improvement (see Figure~\ref{fig:overview}). However, despite substantial progress in human pose understanding, delivering reliable corrective feedback in unconstrained real-world settings remains an open challenge.

Existing methods remain limited in their applicability to practical yoga training. Specifically, many methods emphasize pose recognition or coarse scoring, while providing only limited fine-grained feedback for corrective guidance \cite{ref_article1, ref_article9}. Additionally, classification-based models often struggle to distinguish between visually similar poses, leading to ambiguous or unstable predictions \cite{ref_article15, ref_article17}. Their performance is further challenged by variations in viewpoint, background, and user appearance, reducing robustness in real-world scenarios \cite{ref_article13}. Moreover, most prior works treat pose estimation, recognition, and correction as disjoint tasks, which hinders the delivery of consistent and semantically meaningful guidance \cite{ref_article1}.

\begin{figure}[t]
    \centering
    \includegraphics[width=.9\linewidth]{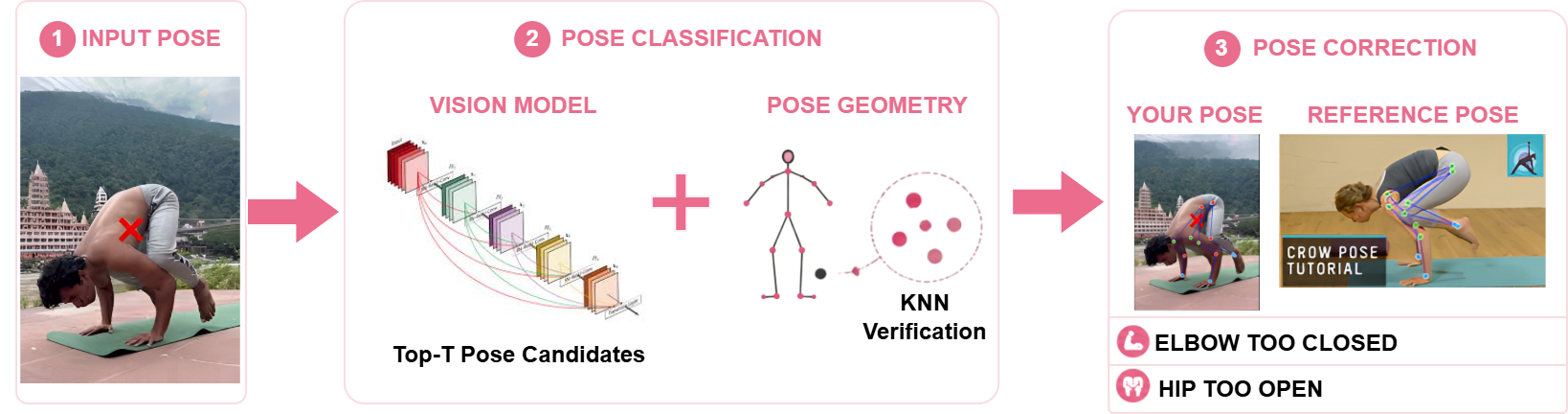}
    \vspace{-1mm}
    \caption{Overview of SAGE-Yoga. Visual models first produce the top-$T$ candidate classes, then MediaPipe landmarks are used for KNN-based geometric verification. The verified pose is compared with a reference to estimate joint-angle deviations and generate corrective feedback.}
    \label{fig:overview}
    \vspace{-7mm}
\end{figure}

To overcome these limitations, we draw inspiration from real-world yoga instruction. Human assessment of yoga poses is rarely based on a single visual cue. Experienced instructors typically consider multiple aspects of a practitioner's posture before deciding whether a pose is correctly performed. This observation motivates our design philosophy: rather than relying on the prediction of a single visual model, multiple complementary models first provide candidate interpretations of the input pose, which are subsequently verified through explicit geometric reasoning. Following this intuition, we propose \textbf{\textit{s}}keletal \textbf{\textit{a}}lignment and \textbf{\textit{g}}eometric \textbf{\textit{e}}nsemble for yoga pose analysis (SAGE-Yoga), a unified framework that jointly addresses pose classification and joint-level correction through a coarse-to-fine pipeline. Given a single RGB image, SAGE-Yoga first applies an ensemble-based classifier to predict candidate pose classes, yielding an \emph{effective} initial estimate over 82 yoga poses. These candidates are then refined through geometric verification based on 2D skeletal keypoints, which makes the framework more \emph{robust} by reducing prediction ambiguity and preserving stable performance across most classes. The verified pose is subsequently compared with a reference pose model to estimate joint-wise deviations and generate corrective instructions. Moreover, this refinement process is \emph{discriminative}, as the proposed KNN-based module better separates visually similar poses and improves the accuracy of the final prediction.

Our contributions are summarized as follows:
\begin{itemize}
\vspace{-1mm}
    \item We formulate pose correction as a reference-based geometric comparison task that translates joint-wise deviations between a detected pose and a retrieved reference pose into interpretable and actionable corrective feedback.
    
    \item We propose SAGE-Yoga, a coarse-to-fine yoga pose classification framework that integrates ensemble-based probabilistic scoring with KNN refinement to accurately distinguish a large set of visually similar yoga poses.
    
    \item We demonstrate that bootstrap bagging over modern convolutional representations substantially improves classification reliability on the Yoga-82 dataset~\cite{ref_article9}, yielding strong performance across diverse pose categories.
\end{itemize}
\vspace{-3mm}
\section{Related Work} \label{relatedworks}
\vspace{-2mm}
In this section, we review prior studies on human pose estimation, yoga pose recognition and assessment, and coarse-to-fine pose recognition to contextualize our proposed approach.

\noindent\textbf{Human Pose Estimation} has evolved from handcrafted features and part-based models~\cite{ref_article5, ref_article6}, which are often sensitive to occlusion and viewpoint variations, to deep learning-based approaches for keypoint detection. For example, OpenPose~\cite{ref_lncs7} predicts body joints using confidence maps and part affinity fields to assemble full-body skeletons, whereas HRNet~\cite{ref_lncs8} preserves high-resolution feature maps through multi-scale fusion to achieve precise localization. MediaPipe~\cite{ref_article2}, which is adopted in this work, uses a lightweight regression-based pipeline for real-time estimation of 33 body keypoints. Although these methods are highly effective at localizing keypoints, they do not inherently assess pose correctness or provide corrective feedback. Our work builds on these foundation to estimate coordinates to enable geometric pose modeling and correction (see Sections \ref{subsec:geometry} and~\ref{subsec:correction}).

\noindent\textbf{Yoga Pose Recognition and Assessment} typically relies on CNN-based architectures, including dual-stream models~\cite{ref_article16}, multi-scale detectors~\cite{ref_article15}, and attention-based ensembles~\cite{ref_article17}, to classify pose categories from images or skeletal keypoints. However, these methods primarily focus on inter-class discrimination and are often less effective at capturing subtle intra-class variations. Complementary pose assessment methods estimate execution quality using contrastive skeletal representations~\cite{ref_article18} or geometric deviations between keypoints~\cite{ref_article19}. Although effective for quality scoring, such methods frequently rely on predefined rules and provide limited explanatory depth, restricting their ability to deliver fine-grained, joint-specific corrective feedback. Our work addresses these limitations through pose classification, reference-based correction, and explicit feedback generation (see Sections \ref{subsec:classification}, \ref{subsec:correction}, and~\ref{subsec:feedback}).

\noindent\textbf{Coarse-to-Fine and Multi-Stage Pose Recognition} has proven effective for pose recognition by combining visual classification with geometric refinement. In particular, such models first generate candidate pose predictions, which are then refined using keypoint-based representations or angle-based likelihoods~\cite{ref_article1, ref_article9}. In addition, multi-stage and ensemble techniques improve robustness through hierarchical classifiers or iterative refinement~\cite{ref_article10, ref_article11}. However, these approaches primarily focus on improving classification accuracy, with limited attention to pose correction. Our work extends this paradigm by integrating ensemble-based classification with geometric verification (see Sections \ref{subsec:ensemble} and~\ref{subsec:geometry}).

\section{The Proposed Approach}\label{sec:approach}
\vspace{-1mm}
\begin{figure}[!t]
    \centering
    \includegraphics[width=.9\linewidth]{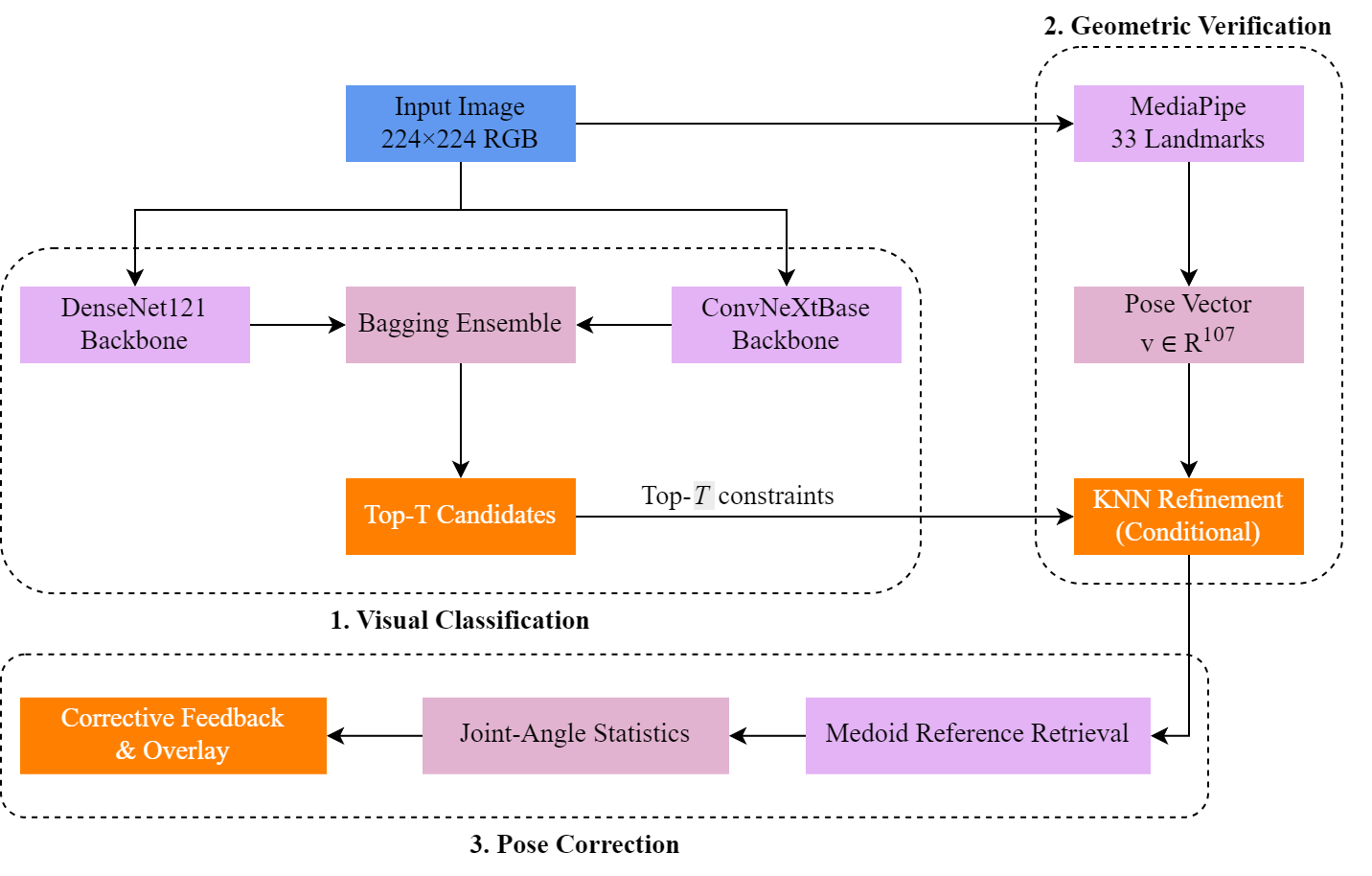}
    \vspace{-1.5mm}
    \caption{Coarse-to-fine SAGE-Yoga architecture. A bagging-based visual ensemble generates Top-$T$ pose candidates, which constrain KNN refinement over geometric features. The final prediction is used for medoid-based joint-level correction and personalized feedback.}
\label{fig:model_architecture}
\vspace{-7mm}
\end{figure}

\noindent\textbf{Overview.} Given an RGB image, SAGE-Yoga performs pose classification, geometric verification, and pose correction in a coarse-to-fine manner. As shown in Figure~\ref{fig:model_architecture}, a bagging-based visual ensemble first generates a Top-$T$ candidate set, which is then refined using a 33-keypoint skeletal representation. The final prediction retrieves a reference pose whose joint-angle statistics guide joint-level correction and feedback generation. Compared with prior single-model pipelines and heuristic correction strategies~\cite{ref_article1}, SAGE-Yoga improves classification robustness, geometric precision, and correction reliability.

\vspace{-1mm}
\subsection{Pose Classification}\label{subsec:classification}
\vspace{-1mm}
This stage aims to identify the yoga pose depicted in the input image based solely on visual appearance. Specifically, we first evaluate six convolutional neural network (CNN) architectures as candidate feature extractors: ResNet50V2~\cite{ref_article21}, MobileNetV2~\cite{ref_article22}, NASNetMobile~\cite{ref_article23}, EfficientNetV2B0~\cite{ref_article24}, DenseNet121~\cite{ref_article25}, and ConvNeXtBase~\cite{ref_article26}. Based on their empirical performance, as reported in Table~\ref{tab:model_performance}, we select DenseNet121~\cite{ref_article25} and ConvNeXtBase~\cite{ref_article26} as the visual backbones of the classification module. Subsequently, these backbones extract high-level visual representations from the input image and generate a ranked list of candidate pose classes. Instead of making the final prediction using a single classifier, we feed the extracted features into a bagging-based ensemble, described in Section~\ref{subsec:ensemble}, which produces a top-$T$ candidate set for subsequent geometric verification in Section~\ref{subsec:geometry}.

To improve fine-tuning stability, we adopt a progressive unfreezing strategy. Specifically, training begins with frozen backbone networks, updating only the newly added classification layers. The backbones are then gradually unfrozen in three stages (i.e., 25\%, 60\%, and 100\%), enabling the pretrained representations to adapt smoothly to yoga pose classification.

\vspace{-1mm}
\subsection{Bagging-Based Ensemble}\label{subsec:ensemble}
\vspace{-1mm}
Given the visual backbones selected in Section~\ref{subsec:classification}, we construct a bagging-based ensemble to improve the robustness of pose prediction. For each input image, DenseNet121 and ConvNeXtBase extract complementary visual descriptors, which are independently normalized using per-sample $L_2$ normalization and then concatenated into a fused 2048-dimensional representation. This fused feature serves as the input to a set of classifier heads with identical architecture. Each head is implemented as a multi-layer perceptron that maps the shared representation to probabilities over the 82 yoga pose classes.

We train the ensemble using stratified bootstrap sampling to encourage diversity among classifier heads while preserving the original class distribution. The main hyperparameters are tuned on the training set and then fixed across all heads so that performance differences primarily reflect bootstrap resampling rather than changes in model configuration. Each MLP head is then trained independently on its corresponding bootstrap sample, and the checkpoint with the lowest validation loss is retained. At inference time, the outputs of the 20 MLP heads are aggregated to obtain the final class-probability distribution, from which the Top-$T$ candidate pose classes are passed to the geometric verification stage described in Section~\ref{subsec:geometry}.

\vspace{-1mm}
\subsection{Geometric Pose Representation and Verification}\label{subsec:geometry}
\vspace{-1mm}
Given the top-$T$ candidate poses generated by the bagging-based ensemble in Section~\ref{subsec:ensemble}, we then refine the prediction using pose geometry. Specifically, we extract human pose landmarks using MediaPipe~\cite{ref_article2}, which provides 33 body keypoints for each image. Each pose is then represented by a 107-dimensional geometric descriptor comprising 99 spatial coordinates $(x,y,z)$ from 33 landmarks and 8 angles.

Additionally, we perform selective verification using a margin-based gating mechanism. If the confidence gap between the top-1 and top-2 ensemble predictions exceeds a predefined threshold, the visual prediction is accepted directly. Otherwise, the sample is considered ambiguous, and KNN verification is applied within the top-$T$ candidate set. To obtain a balanced similarity measure, the spatial and angular components are compared separately, with distances in each subspace normalized via min-max scaling. The final distance is defined in Eq.~\eqref{eq:angular}.
\begin{equation}
    D_{total} = \alpha \cdot \tilde{D}_{spatial} + (1 - \alpha) \cdot \tilde{D}_{angular}
    \label{eq:angular}
\end{equation}
where $\alpha$ controls the relative contributions of spatial and angular information. Its influence is systematically evaluated in the ablation study (Figure~\ref{fig:margin_alpha_grid}), which shows that SAGE-Yoga maintains stable performance across a broad range of $\alpha$ values. Based on this analysis, we adopt $\alpha=0.45$ for all experiments. The final pose label is determined by majority voting over the $K$ nearest neighbors under this fused distance. The sensitivity of the proposed refinement to the choice of $K$ and the candidate pool size $T$ is analyzed in Section~\ref{subsubsec:knn_candidates}. This voting scheme improves robustness by aggregating local neighborhood information, but may introduce ambiguity when the top two candidate classes receive the same number of votes. The frequency of such cases depends on both the candidate set size $T$ and the number of neighbors $K$, as analyzed in Figure~\ref{fig:knn_tie_count}. In these cases, the ambiguity is resolved by selecting the class with the smaller average distance among its neighbors.
\begin{figure}[!t]
    \centering
    \includegraphics[width=.9\linewidth]{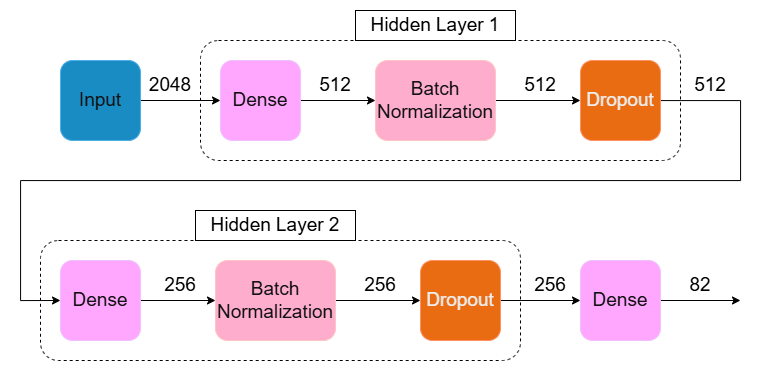}
    \vspace{-2mm}
    \caption{MLP classification head. The fused 2048-dimensional features are compressed by two fully connected layers with batch normalization and dropout, followed by a softmax classifier over 82 yoga poses.}
    \label{fig:mlp}
    \vspace{-7mm}
\end{figure}
\vspace{-1mm}
\subsection{Medoid-Based Reference Pose Correction}\label{subsec:correction}
\vspace{-1mm}
Given the final pose label obtained after geometric verification in Section~\ref{subsec:geometry}, we perform pose correction by comparing the user's pose with a representative reference pose from the predicted class. For each yoga pose class, the reference is selected automatically from the training set using a medoid-based strategy in joint-angle space. Unlike a centroid or mean representation, the medoid corresponds to an actual training sample whose joint-angle configuration minimizes the total distance to all other samples in the same class, yielding a reference pose that is both typical and physically realizable.
To account for natural variations in pose execution, we model the class-specific distribution of selected joint angles by computing their standard deviations. During inference, the user's joint angles are compared with those of the medoid reference and evaluated against these learned statistics. A joint is considered misaligned if its angle falls outside the acceptable range defined by the class distribution. The detected deviations are then used to generate targeted corrective feedback.

\vspace{-1mm}
\subsection{Pose Correction Feedback}\label{subsec:feedback}
\vspace{-1mm}
After the final pose class is predicted, SAGE-Yoga generates feedback using a rule-based text template. First, the model reports the classified pose. Next, it compares each evaluated joint angle with the acceptable angle range learned from reference samples of the predicted class. When a joint angle falls below this range, the joint is considered too closed, and the system suggests opening or straightening it. Conversely, when a joint angle exceeds the range, the joint is considered too open, and the system suggests relaxing or bending it.

When misaligned joints are detected, the system presents structured feedback showing the classified pose, followed by ``Potential Pose Corrections Found''. Each correction identifies the misaligned joint, its current angle, the recommended action, and the target angle range. If no correction is needed, the system outputs a positive confirmation, such as ``Excellent! Your \texttt{<class>} posture matches the reference pose.''

\vspace{-1mm}
\subsection{Loss Function}\label{subsec:loss}
\vspace{-1mm}
In our proposed method, Categorical Cross-Entropy (CCE) is a loss function used to measure the difference between the actual labels and the predicted labels. This loss function is suitable for SAGE-Yoga because it is commonly used for multi-class classification tasks such as 82 yoga postures \cite{ref_article9}, \cite{ref_article15}, \cite{ref_article17}. For a single sample, the formula is formally defined in Eq.~\eqref{eq:lossfunction}.
\begin{equation}
    L = - \sum^{82}_{i=1}y_ilog(\hat{y_i})
    \label{eq:lossfunction}
\end{equation}
where $y$ is the one-hot ground-truth label and $\hat{y}$ is the predicted class-probability vector. The CCE loss penalizes low probability assigned to the correct class, so training aims to minimize loss.


\section{Experiment Results}\label{sec:experiment}
\vspace{-1mm}
\begin{figure}[!t]
    \centering
    \begin{minipage}[c]{0.45\linewidth}
        \centering
        \begin{tabular}{|c|l|} 
            \hline
            \textbf{No.} & \textbf{Joint Angles} \\ \hline
            1 & Left Elbow \\ \hline
            2 & Left Shoulder  \\ \hline
            3 & Right Shoulder \\ \hline
            4 & Right Elbow \\ \hline
            5 & Left Hip \\ \hline
            6 & Right Hip \\ \hline
            7 & Left Knee \\ \hline
            8 & Right Knee \\ \hline
        \end{tabular}
    \end{minipage}
    \hfill 
    \begin{minipage}[c]{0.5\linewidth}
        \centering
        \includegraphics[width=\linewidth]{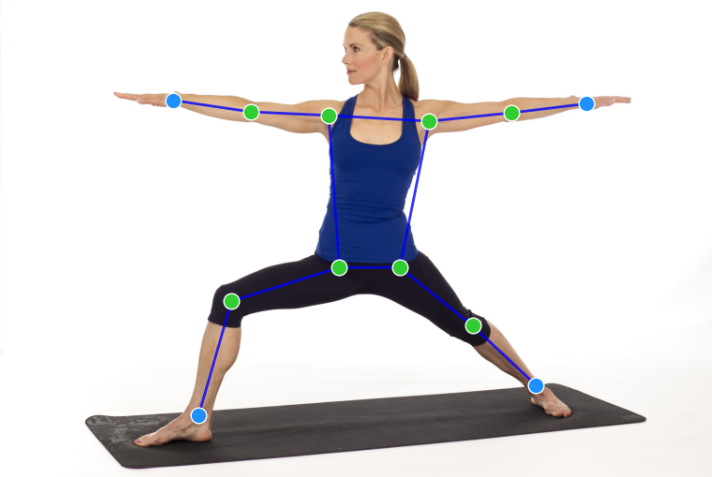}
    \end{minipage}
    \caption{Warrior II reference pose from Yoga-82 with eight evaluated joint angles computed from MediaPipe keypoints. The angle vector provides a compact representation for classification and joint-wise correction. \textcolor{green}{Green} marks evaluated aligned joints; \textcolor{blue}{blue} marks joints not used for correction.}
    \label{fig:warrior_angles}
    \vspace{-7mm}
\end{figure}

\subsection{Experimental Setup}
\vspace{-1mm}
\noindent\textbf{Model Configuration.}
We first convert each input image to RGB and resize it to \(224 \times 224\) pixels to match the input resolution of the backbone networks. We then employ a dual-backbone feature extraction framework followed by a lightweight MLP classifier. During training, the convolutional backbones are fine-tuned using the progressive unfreezing strategy described in Section~\ref{subsec:classification}. Training is performed for up to 100 epochs using categorical cross-entropy loss and the \texttt{Adam} optimizer, with an initial learning rate of \(10^{-3}\), a batch size of 64, and early stopping based on validation loss.

All experiments are conducted on an NVIDIA Tesla P100 PCIe GPU with 16 GB of memory and 31.35 GB of system RAM for CPU-bound preprocessing and data loading.

\noindent\textbf{Dataset.}
We evaluate SAGE-Yoga on the Yoga-82 dataset~\cite{ref_article9}, which contains 16,767 RGB images spanning 82 pose classes under diverse conditions, including variations in viewpoint, illumination, background, and subject appearance. The dataset is divided into 11,652 training images, 3,351 validation images, and 1,764 test images, with notable class imbalance ranging from 35 to 499 images per class (mean: 204.5; median: 192.5). Additionally, we apply augmentation, expanding the training set to 34,956 images, while the validation and test sets remain unchanged. All images are resized to $224 \times 224$ pixels.

To preserve pose geometry while improving generalization, we use a lightweight preprocessing pipeline comprising RGB conversion and architecture-specific normalization. During training, we apply pose-preserving augmentations with Albumentations, including horizontal flips, brightness/contrast and HSV perturbations, affine transformations with moderate shift, scale, and rotation, and random resized cropping. We keep the augmentation strength moderate to avoid disrupting joint relationships while improving robustness to visual variation. 

To obtain a compact geometric representation, we further compute joint angles from selected MediaPipe keypoints, as shown in Figure~\ref{fig:warrior_angles}. The eight evaluated joints correspond to key body articulations and provide an expressive representation of pose structure for downstream error detection.

\noindent\textbf{Metrics.} We evaluate classification performance using Top-1 accuracy, which measures the proportion of pose labels predicted correctly. Additionally, we report Top-$K$ accuracy.

\vspace{-1mm}
\subsection{Quantitative Results}\label{subsec:quant}
\vspace{-1mm}


\begin{table}[t]
\centering
\caption{
Comparison on the Yoga-82 dataset. $\lambda$ denotes the baseline-specific KNN coefficient, $m$ is the margin threshold for selective refinement, $T$ is the top-$T$ candidate pool size, $K$ is the number of KNN voters, and $\alpha$ is the spatial-angular fusion weight in Eq.~\eqref{eq:angular}.
}
\label{tab:final_model_performance}
\resizebox{\columnwidth}{!}{%
\begin{tabular}{lll}
\toprule
\textbf{Method} & \textbf{Top-1 Acc.} & \textbf{Setting} \\
\midrule
KNN & 0.72 & $\lambda=9,\ K=4$ \\
DenseNet121 + ConvNeXtBase & 0.89 & visual ensemble only \\
\textbf{SAGE-Yoga (Ours)} & \textbf{0.91} & $m=0.8,\ T=3,\ K=2,\ \alpha=0.45$ \\
\bottomrule
\end{tabular}}
\end{table}

\noindent\textbf{Classification Performance.}
Table~\ref{tab:final_model_performance} reports the top-1 accuracy on Yoga-82. The KNN achieves an accuracy of only 0.72, suggesting that pose geometry alone is insufficient for fine-grained recognition. In contrast, the DenseNet121+ConvNeXtBase ensemble improves the accuracy to 0.89, indicating that deep visual representations provide the dominant classification signal. This backbone choice is supported by Section~\ref{subsubsec:backbone_selection}, where ConvNeXtBase-based ensembles obtain the strongest top-$k$ results, while the combination of DenseNet121 and ConvNeXtBase delivers the most consistent performance.

SAGE-Yoga further improves the accuracy to 0.91 through selective geometric verification. Specifically, the visual ensemble first generates a Top-$T$ candidate set, after which KNN refinement is applied only when the prediction margin falls below $m$. This refinement uses a fused spatial-angular distance weighted by $\alpha$. The selected settings, $m=0.8$ and $\alpha=0.45$, are supported by Section~\ref{subsubsec:margin_fusion}, while $T=3$ and $K=2$ are supported by Section~\ref{subsubsec:knn_candidates}. Overall, the improvement over the ensemble baseline suggests that geometry is most effective for resolving ambiguous visual predictions. This conclusion is further validated by Section~\ref{subsubsec:agreement_analysis}, where KNN refinement corrects substantially more ensemble errors than it introduces.

\begin{table}[t]
\vspace{-2mm}
\centering
\caption{Comparison between the Yoga-82 baseline variants and the proposed SAGE-Yoga. Top-1 accuracy and Macro-F1 are reported to evaluate overall classification performance and robustness to class imbalance.}
\label{tab:model_comparison}
\renewcommand{\arraystretch}{1.15}
\begin{tabular}{lccc}
\hline
\textbf{Method} & \textbf{Parameters} & \textbf{Top-1 Acc.} & \textbf{Macro-F1} \\
\hline
Yoga-82 Variant 1 & 18.27 M & 0.7607 & 0.7366 \\
Yoga-82 Variant 2 & 18,58 M & 0.7333 & 0.7076 \\
Yoga-82 Variant 3 & 22,60 M & 0.7430 & 0.7085 \\
\hline
\textbf{SAGE-Yoga (Ours)} & - M & \textbf{0.9070} & \textbf{0.9011} \\
\hline
\end{tabular}
\vspace{-5mm}
\end{table}

\noindent\textbf{Comparison with Baseline Models.}
Table~\ref{tab:model_comparison} compares SAGE-Yoga with Yoga-82 variants~\cite{ref_article9} that differ in the placement and capacity of auxiliary classifiers for the three-level label hierarchy. For a fair comparison, we re-implemented all baselines using the authors' released code and original training settings and evaluated them under the same protocol. Variant~1, which attaches the level-1 and level-2 classifiers after DenseBlocks~2 and~3, respectively, achieves better Top-1 accuracy and Macro-F1 than Variant~2, suggesting that earlier supervision better preserves coarse pose structure. It also outperforms Variant~3 despite using fewer parameters, indicating that increasing the capacity of the coarse-level branch does not improve fine-grained recognition. SAGE-Yoga achieves the best overall performance, with a Top-1 accuracy of 0.9070 and a Macro-F1 score of 0.9011, substantially surpassing all baselines. The strong Macro-F1 result further demonstrates robust performance across pose categories under class imbalance.

\subsection{Qualitative Pose Correction}\label{subsec:qual}
\vspace{-1mm}
\begin{figure}[!t]
    \centering
    \begin{tabular}{cl}
        \rotatebox{90}{\footnotesize \textbf{Yoga-82}} & 
        
        \begin{minipage}{0.92\linewidth} 
            \centering
            \includegraphics[width=0.23\linewidth]{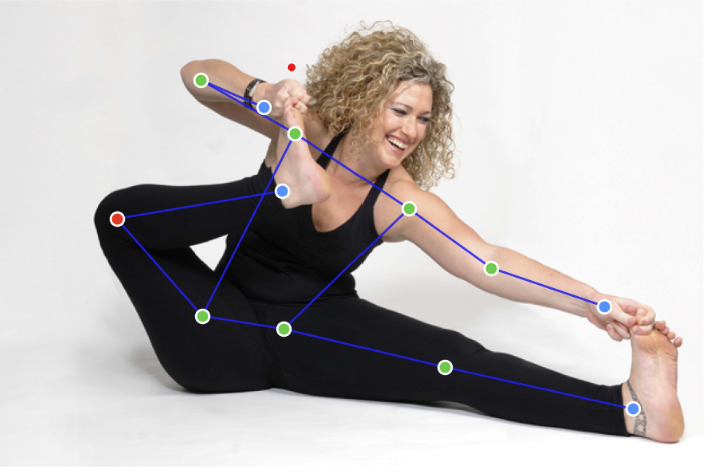} \hfill
            \includegraphics[width=0.23\linewidth]{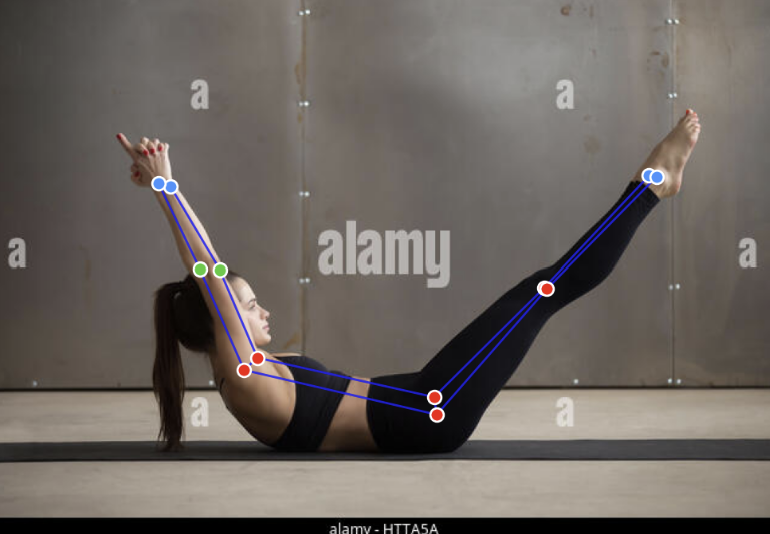} \hfill
            \includegraphics[width=0.23\linewidth]{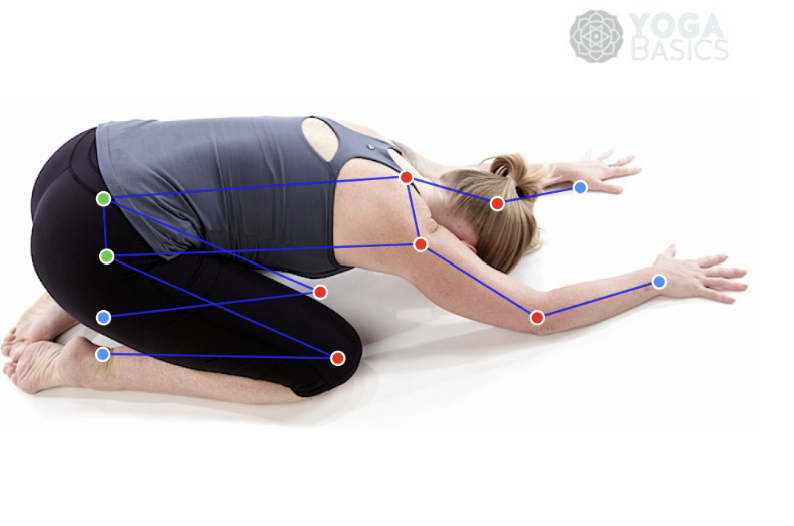} \hfill
            \includegraphics[width=0.23\linewidth]{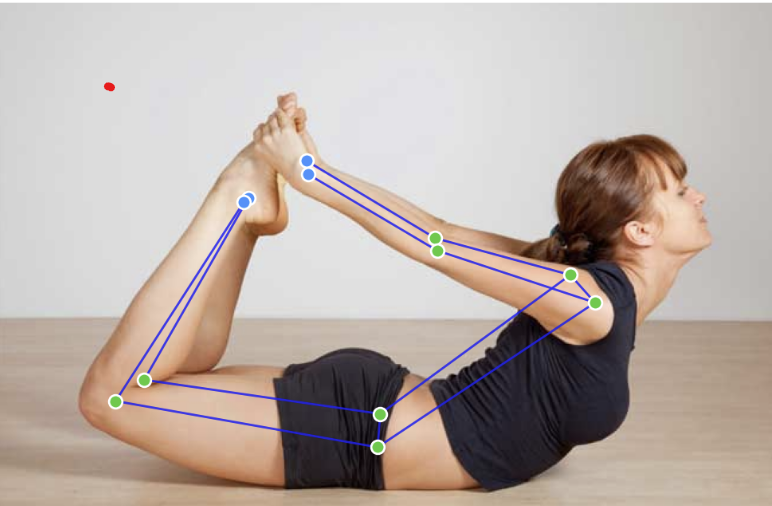} \\
            
            \vspace{0.05cm} 
            
            \includegraphics[width=0.23\linewidth]{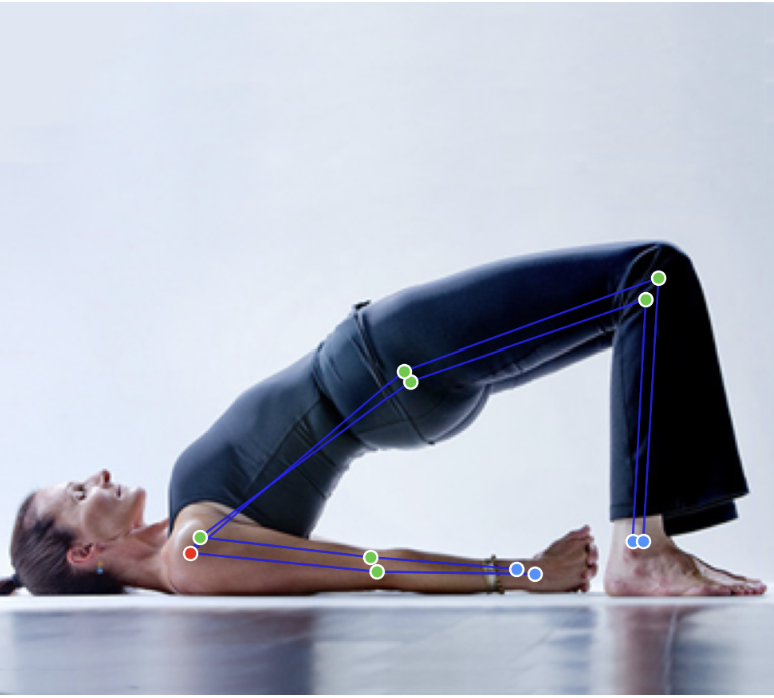} \hfill
            \includegraphics[width=0.23\linewidth]{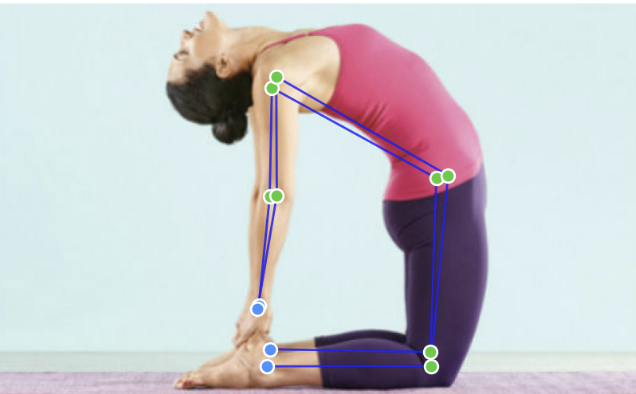} \hfill
            \includegraphics[width=0.23\linewidth]{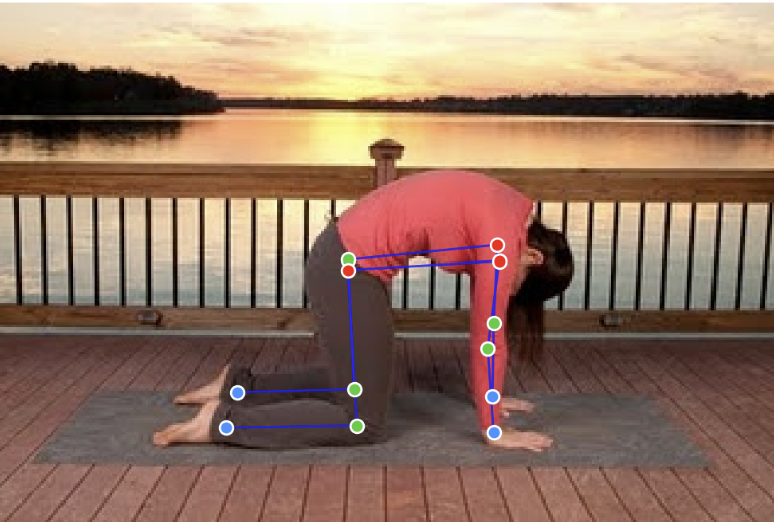} \hfill
            \includegraphics[width=0.23\linewidth]{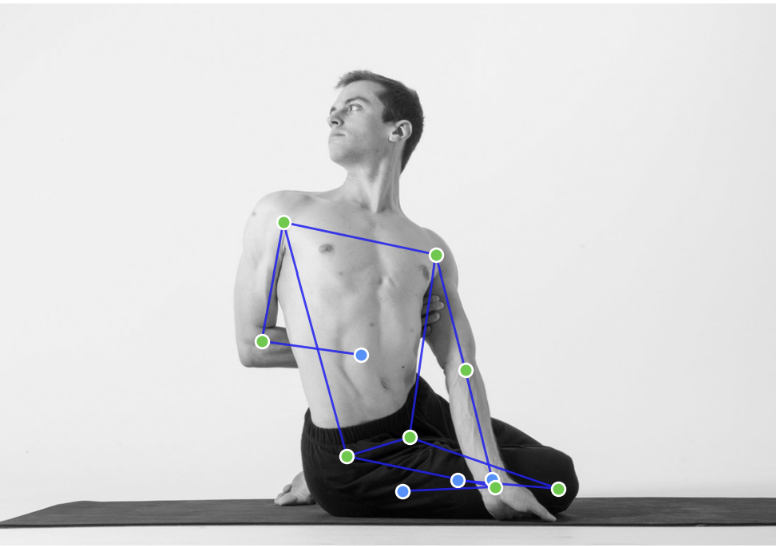}
        \end{minipage}
    \end{tabular}
    \vspace{-0.25mm}
    \caption{Qualitative SAGE-Yoga predictions with skeletal keypoints. \textcolor{green}{Green} denotes aligned joints, \textcolor{red}{red} indicates significant deviations, and \textcolor{blue}{blue} marks joints not evaluated for correction.}
    \label{fig:yoga82_samples}
    \vspace{-5mm}
\end{figure}

The qualitative examples in Figure~\ref{fig:yoga82_samples} demonstrate that SAGE-Yoga provides both pose-level classification and joint-level interpretability. Rather than merely predicting a yoga pose label, the model also identifies local body regions that deviate from the learned pose distribution. This capability is important for yoga analysis, since two samples may belong to the same class while differing substantially in execution quality.

Furthermore, Figure~\ref{fig:pose_correction} illustrates the complete correction output for a sample classified as \texttt{Staff\_Pose\_or\_Dandasana}. Specifically, the model identifies elbows as overly closed relative to the class-specific reference ranges and generates corrective instructions to open or straighten these joints. Notably, the visual and textual outputs are consistent, as the red-highlighted joints correspond to the body parts described in the generated feedback.

\subsection{Ablation Study}
\label{subsec:ablation}
\vspace{-1mm}
\begin{figure}[t]
    \centering
    \begin{minipage}[c]{0.65\linewidth}
    1. Classified Pose: 
    
    Staff\_Pose\_or\_Dandasana\_
    
    2. Geometric Feedback:
    
    Potential Pose Corrections Found:
    \begin{adjustwidth}{2em}{0pt}
    \begin{itemize}
        \item \textbf{LEFT ELBOW} is too \textbf{closed} (80.9°). Try to \textbf{open/straighten} this joint (Target: 120.6° - 198.0°).
        \item \textbf{RIGHT ELBOW} is too \textbf{closed} (90.9°). Try to \textbf{open/straighten} this joint (Target: 125.1° - 199.6°).
    \end{itemize}
    \end{adjustwidth}
    \end{minipage}
    \hfill 
    \begin{minipage}[c]{0.3\linewidth}
        \centering
            \includegraphics[width=0.8\linewidth]{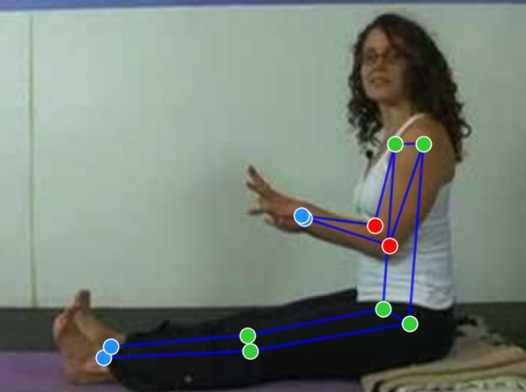}
            \includegraphics[width=0.8\linewidth]{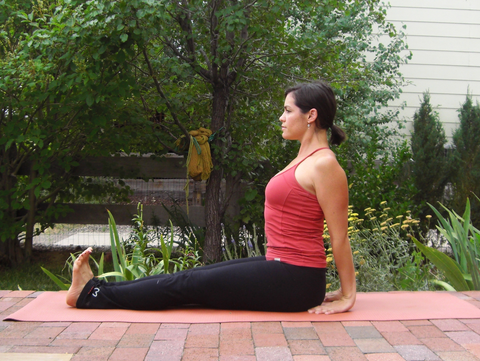}
    \end{minipage}
    \vspace{1mm}
    \caption{SAGE-Yoga provides visual and textual feedback. \textcolor{green}{Green} marks aligned joints, \textcolor{red}{red} marks deviations, and \textcolor{blue}{blue} marks joints not evaluated.}
    \label{fig:pose_correction}
    \vspace{-5mm}
\end{figure}

\subsubsection{Margin-Gated Spatial--Angular Refinement}
\label{subsubsec:margin_fusion}
\begin{figure}[!t]
    \centering
    \vspace{-5mm}
    \includegraphics[width=.85\linewidth]{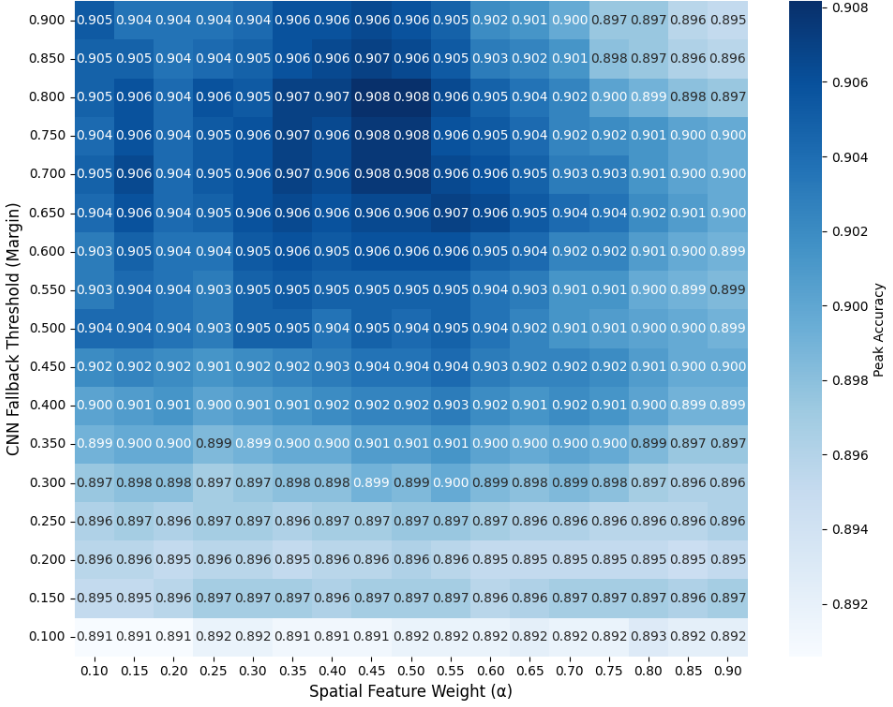}
    \caption{Grid search over margin threshold $m$ and spatial weight $\alpha$. Performance peaks near $m=0.8$ and $\alpha=0.45$, indicating that SAGE-Yoga benefits from selective refinement and balanced spatial--angular geometry.}
    \label{fig:margin_alpha_grid}
    \vspace{-2mm}
\end{figure}

Figure~\ref{fig:margin_alpha_grid} analyzes the sensitivity of the geometric verification stage described in Section~\ref{subsec:geometry} to the margin threshold $m$, which determines when the visual prediction is retained, and the weighting factor $\alpha$, which balances the normalized spatial and angular distances. Accuracy remains above 0.90 across a broad central region, indicating that the model is not narrowly tuned to a specific parameter setting. The best performance occurs at $m\in[0.75,0.85]$ and $\alpha\in[0.35,0.55]$, with $m=0.8$ and $\alpha=0.45$ selected for the final model. These results indicate that geometric refinement is beneficial for uncertain predictions and that a balanced combination of spatial and angular cues, consistent with the 107-dimensional pose descriptor defined in Section~\ref{subsec:geometry}, outperforms either cue alone.



\subsubsection{Visual Backbone and Feature-Fusion Selection}
\label{subsubsec:backbone_selection}
\begin{table}[t]
\centering
\caption{
Top-$k$ accuracy of individual backbones and two-model visual ensembles on Yoga-82. ConvNeXtBase-based ensembles achieve the strongest overall performance. We adopt DenseNet121+ConvNeXtBase because it obtains the best Top-1 accuracy and consistently strong Top-$k$ results.
}
\label{tab:model_performance}
\resizebox{.9\columnwidth}{!}{%
\begin{tabular}{lccccc}
\toprule
\textbf{Model} & \textbf{Top-1} & \textbf{Top-2} & \textbf{Top-3} & \textbf{Top-4} & \textbf{Top-5} \\
\midrule
ResNet50V2 & 0.77 & 0.87 & 0.91 & 0.93 & 0.95 \\
MobileNetV2 & 0.79 & 0.88 & 0.92 & 0.94 & 0.95 \\
ResNet50V2 + MobileNetV2 & 0.80 & 0.88 & 0.91 & 0.93 & 0.95 \\
DenseNet121 & 0.84 & 0.91 & 0.94 & 0.96 & 0.97 \\
ResNet50V2 + DenseNet121 & 0.81 & 0.89 & 0.93 & 0.95 & 0.96 \\
MobileNetV2 + DenseNet121 & 0.84 & 0.92 & 0.94 & 0.96 & 0.96 \\
NASNetMobile & 0.75 & 0.86 & 0.90 & 0.92 & 0.94 \\
ResNet50V2 + NASNetMobile & 0.79 & 0.88 & 0.91 & 0.94 & 0.95 \\
MobileNetV2 + NASNetMobile & 0.81 & 0.89 & 0.92 & 0.94 & 0.96 \\
DenseNet121 + NASNetMobile & 0.83 & 0.92 & 0.94 & 0.95 & 0.96 \\
EfficientNetV2B0 & 0.85 & 0.93 & 0.95 & \textbf{0.97} & 0.97 \\
ResNet50V2 + EfficientNetV2B0 & 0.82 & 0.90 & 0.93 & 0.95 & 0.96 \\
MobileNetV2 + EfficientNetV2B0 & 0.84 & 0.92 & 0.95 & 0.96 & 0.97 \\
DenseNet121 + EfficientNetV2B0 & 0.85 & 0.92 & 0.95 & 0.96 & 0.97 \\
NASNetMobile + EfficientNetV2B0 & 0.84 & 0.93 & 0.95 & 0.96 & 0.97 \\
ConvNeXtBase & 0.88 & 0.94 & 0.96 & \textbf{0.97} & \textbf{0.98} \\
ResNet50V2 + ConvNeXtBase & 0.86 & 0.93 & 0.95 & \textbf{0.97} & 0.97 \\
MobileNetV2 + ConvNeXtBase & 0.87 & 0.94 & 0.96 & \textbf{0.97} & 0.97 \\
DenseNet121 + ConvNeXtBase & \textbf{0.89} & \textbf{0.95} & \textbf{0.97} & \textbf{0.97} & \textbf{0.98} \\
NASNetMobile + ConvNeXtBase & \textbf{0.89} & \textbf{0.95} & 0.96 & \textbf{0.97} & 0.97 \\
EfficientNetV2B0 + ConvNeXtBase & \textbf{0.89} & 0.94 & 0.96 & \textbf{0.97} & \textbf{0.98} \\
\bottomrule
\end{tabular}}
\vspace{-7mm}
\end{table}
Table~\ref{tab:model_performance} validates the backbone selection described in Section~\ref{subsec:classification}. ConvNeXtBase achieves the highest individual Top-1 accuracy of 0.88, while DenseNet121 provides competitive Top-$k$ performance. Their combination further improves Top-1 accuracy to 0.89 and yields the best Top-2, Top-3, and Top-5 results, indicating complementary representations: ConvNeXtBase strengthens global recognition, whereas DenseNet121 improves candidate ranking. This finding supports the ensemble design in Section~\ref{subsec:ensemble}, as the geometric verifier operates on the Top-$T$ candidate set and therefore benefits from high Top-$k$ accuracy. Thus, DenseNet121+ConvNeXtBase is selected for its strong Top-1 performance and its consistent inclusion of the correct pose among the candidates passed to geometric refinement.

\subsubsection{Candidate Pool Size, KNN Voter Count, and Voting Ambiguity}
\label{subsubsec:knn_candidates}

\begin{figure}[t]
    \centering
    \includegraphics[width=.9\linewidth]{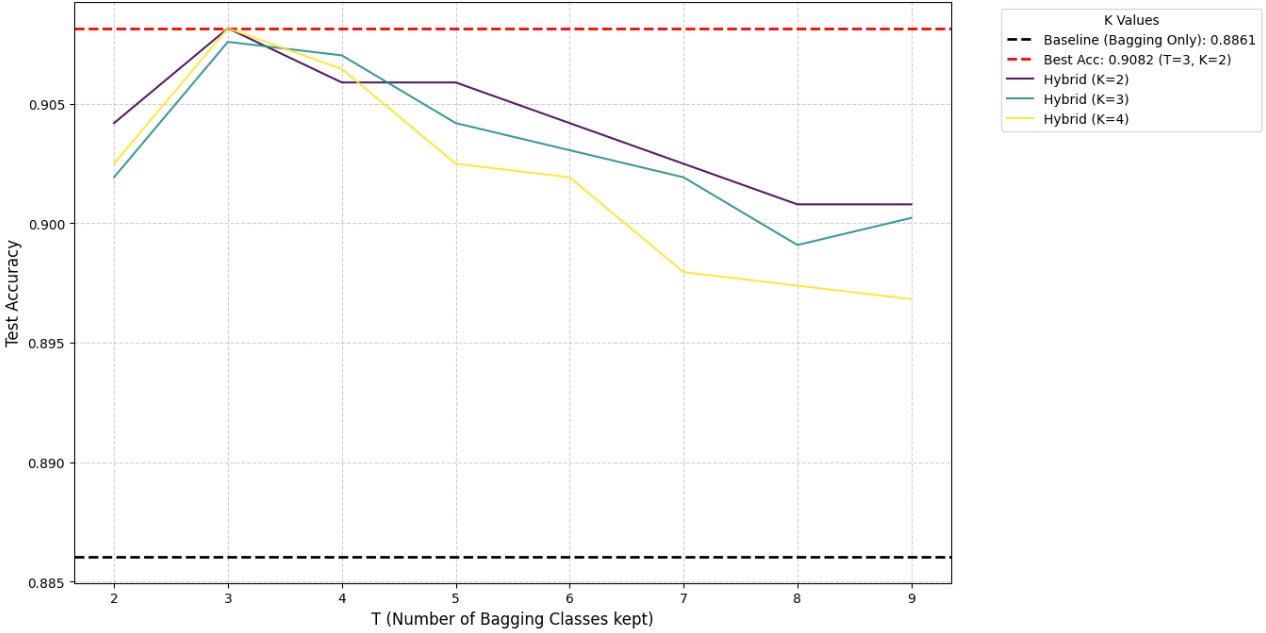}
    \caption{
    Effect of candidate pool size $T$ and KNN voter count $K$ on hybrid classification accuracy. The best performance with $T=3$ and $K=2$.
    }
    \label{fig:knn_accuracy}
    \vspace{-4mm}
\end{figure}
Figure~\ref{fig:knn_accuracy} evaluates the Top-$T$ candidate restriction and KNN voting strategy introduced in Section~\ref{subsec:geometry}. Compared with the bagging-only ensemble accuracy of 0.8861, the best hybrid configuration achieves 0.9082, consistent with the 0.91 accuracy reported in Table~\ref{tab:final_model_performance}. The optimal setting, $T=3$ and $K=2$, supports the coarse-to-fine design of SAGE-Yoga, where visual predictions constrain the search space before local geometric refinement. Increasing $T$ introduces less reliable candidates, whereas increasing $K$ may suppress fine-grained geometric differences. Figure~\ref{fig:knn_tie_count} further analyzes the voting ties described in Section~\ref{subsubsec:knn_candidates}, which are resolved using the smallest average fused distance defined in Section~\ref{subsec:geometry}. Tie frequency increases with $T$ due to stronger inter-class competition and decreases with $K$ through broader neighbor aggregation, explaining why $T=3$ and $K=2$ provide the best balance between low ambiguity and local geometric sensitivity.




\begin{figure}[t]
    \centering
    \includegraphics[width=.75\linewidth]{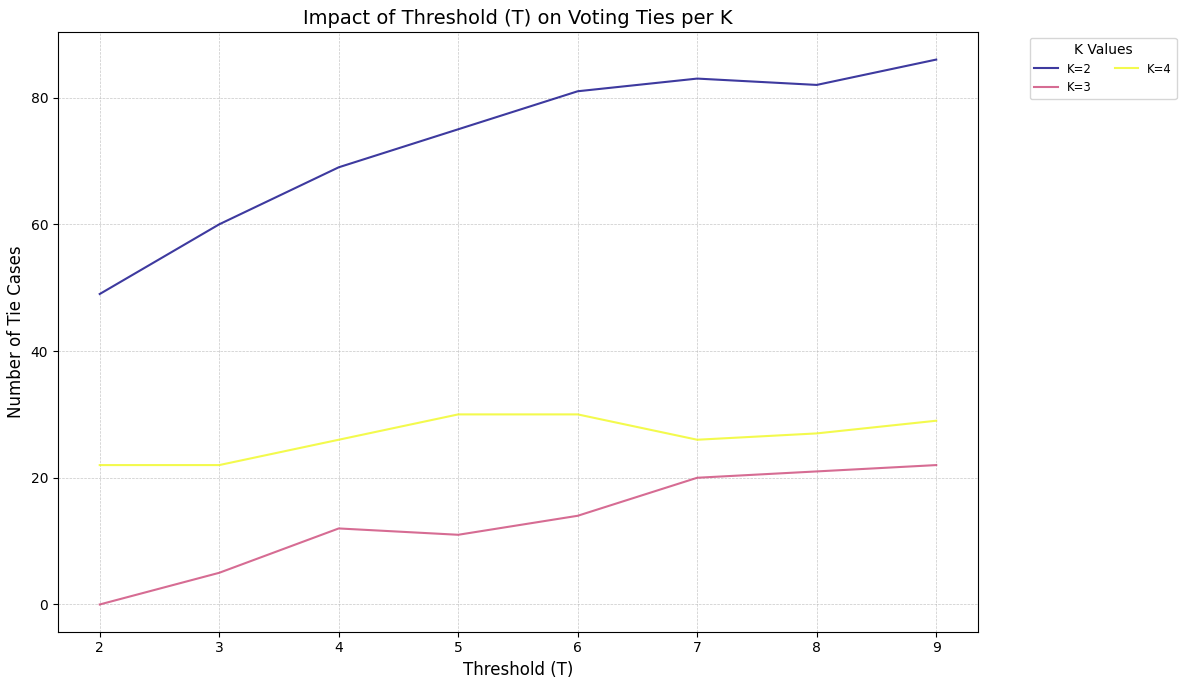}
    \caption{Number of KNN voting ties for different candidate pool sizes $T$ and voter counts $K$. Larger $T$ increases tie frequency by adding competing classes, whereas larger $K$ reduces ties but may weaken local discriminability.}
    \label{fig:knn_tie_count}
    \vspace{-6mm}
\end{figure}



\subsubsection{Contribution of Geometric Refinement}
\label{subsubsec:agreement_analysis}

\begin{figure}[t]
    \centering
    \includegraphics[width=.7\linewidth]{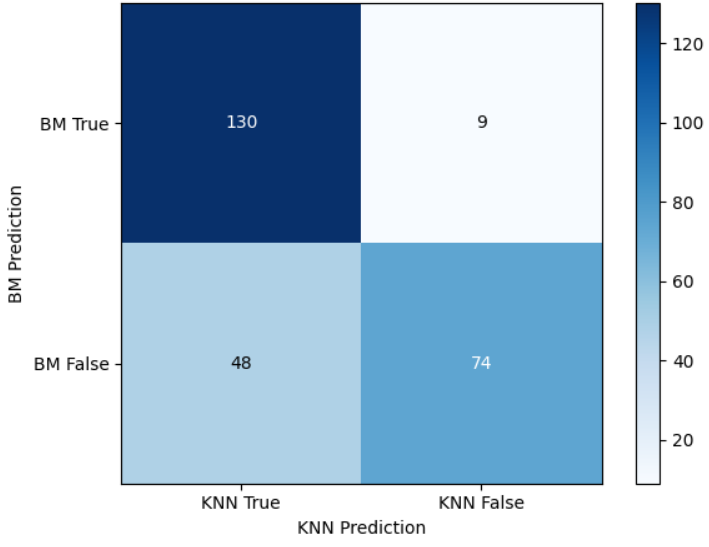}
    \caption{Agreement between the visual ensemble and KNN refinement. Asymmetric disagreement shows KNN improves more cases than it degrades.}
    \label{fig:bagging_knn_agreement}
    \vspace{-7mm}
\end{figure}

Figure~\ref{fig:bagging_knn_agreement} examines the interaction between the geometric verifier and the visual ensemble. Most samples lie on the agreement diagonal, indicating strong consistency between the two modules, while KNN corrects 48 visual errors and introduces only 9 degradations. This asymmetry supports the margin-gated strategy described in Section~\ref{subsec:geometry}: confident visual predictions are retained, whereas geometric refinement is applied primarily to ambiguous cases. Consequently, the hybrid model benefits from complementary geometric cues while avoiding the performance degradation associated with indiscriminate KNN refinement.

\vspace{-1mm}
\section{Conclusion \& Future Work}\label{sec:conclusion}
\vspace{-1mm}
In this paper, we introduced a unified framework for image-based yoga pose analysis that integrates pose recognition and posture correction. Empirically, our extensive experiments on Yoga-82 show that ensemble learning improves top-$K$ recognition performance. We further show that comparing detected poses with class-specific references enables interpretable feedback without temporal cues or additional sensors. These results demonstrate the feasibility of single-image, explainable yoga coaching. Future work will explore adaptive correction, richer feedback, and language-driven personalization.
\vspace{-1.75mm}
{\small
\bibliographystyle{ieeetr}
\bibliography{bibliography}
}
\end{document}